\documentclass{article}

\usepackage{spconf,amsmath,graphicx,enumitem,xcolor,booktabs}
\usepackage[ruled]{algorithm2e}
\usepackage{url}

\title{Robust Marine Buoy Placement for Ship Detection\\ Using Dropout K-Means}

\name{Yuting Ng$^{\star}$ \qquad 
Jo\~ao M. Pereira$^{\star}$ \qquad 
Denis Garagic$^{\dagger}$ \qquad 
Vahid Tarokh$^{\star}$\thanks{This work was supported in part by DARPA Grant No.HR00111990016.}}

\address{$^{\star}$ Duke University \qquad $^{\dagger}$ BAE Systems FAST Labs\textsuperscript{TM}}

\begin{document}

\graphicspath{{./Pictures/}}

\maketitle
\begin{abstract}
Marine buoys aid in the battle against Illegal, Unreported and Unregulated (IUU) fishing by detecting fishing vessels in their vicinity. Marine buoys, however, may be disrupted by natural causes and buoy vandalism. In this paper, we formulate marine buoy placement as a clustering problem, and propose dropout k-means and dropout k-median to improve placement robustness to buoy disruption.

We simulated the passage of ships in the Gabonese waters near West Africa using historical Automatic Identification System (AIS) data, then compared the ship detection probability of dropout k-means to classic k-means and dropout k-median to classic k-median. With 5 buoys, the buoy arrangement computed by classic k-means, dropout k-means, classic k-median and dropout k-median have ship detection probabilities of 38\%, 45\%, 48\% and 52\%. 

\end{abstract}
\section{Introduction}

Illegal, Unreported  and Unregulated (IUU) fishing not only endangers marine ecosystems, but also is a global threat to economic and food security, with annual damages estimated at \$10-23.5 billion and 11-26 million tons~\cite{agnew2009iuu}.  Developing countries which depend on fishing for food and export, such as those in West Africa, are most at risk~\cite{noaa_iuu}. In the battle against IUU fishing, a network of marine buoys can improve the monitoring of fishing activity via ship detection~\cite{bruno2010_satelliteandacoustic, fefilatyev2012_buoycam, prasad2017_buoycam}. 

In May 2013, two marine buoys captured images of a fishing vessel that was fishing illegally. The vessel was caught despite its efforts to conceal its location, by not sending Automatic Identification System (AIS) location reports for a two-week period from May 20 to June 1, 2013~\cite{beets2014_buoyvandalism}. The catch demonstrated the effectiveness of using marine buoys to detect fishing vessels, and there are plans to fit more buoys with cameras~\cite{beets2014_buoyvandalism}. Marine buoys, however, may be disrupted~\cite{teng2009_buoyvandalism}. 

\begin{figure}
    \centering
    \includegraphics[width=\columnwidth]{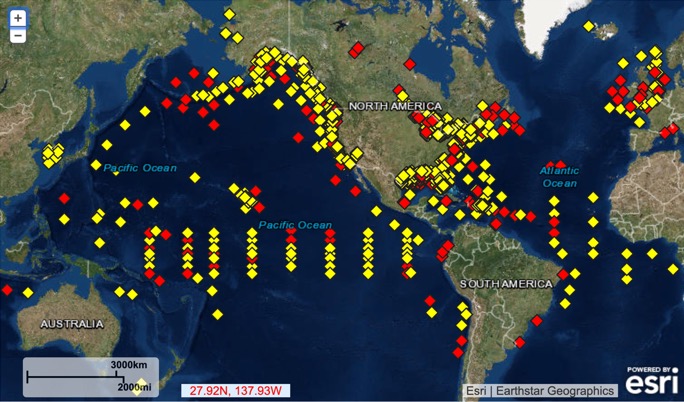}
    \caption{Marine buoys partnered with the National Oceanic and Atmospheric Administration's (NOAA) National Data Buoy Center (NDBC), screenshot taken on July 10 2019~\cite{noaa_ndbc}. Diamonds indicate the locations of buoys: yellow diamonds reported back to NDBC, while red diamonds did not.}
    \label{fig:noaa_ndbc}
\end{figure}

At a given time, only about 70\% of the marine buoys partnered with the National Oceanic and Atmospheric Administration's (NOAA) National Data Buoy Center (NDBC) are reporting. For example, on July 10 2019, out of 1416 buoys deployed, only 1001 buoys reported back to NDBC in an 8~hour period, from 10:30 to 18:30 PDT, as shown in~\figurename~\ref{fig:noaa_ndbc}~\cite{noaa_ndbc}. Communications with buoys are disrupted due to natural causes, such as harsh weather, corrosion, fish bites and marine growth, or acts of vandalism~\cite{beets2014_buoyvandalism}. Fishing vessels engaging in IUU fishing are the most common perpetrators, despite regulations against buoy vandalism~\cite{dbcp2011_buoyvandalism}. 

In order to place buoys, we use a k-means type clustering algorithm to cluster ship positions, and place buoys at the cluster centers obtained by the algorithm. However, buoy disruption changes the spatial configuration of the buoy network. While the ship detection radius of each buoy does not change, the distance of each ship to the closest remaining buoy may increase, and the probability a ship is detected by the buoy network decreases. We thus propose a more robust buoy placement by considering weights proportional to the probability of buoy dropout in the k-means and k-median clustering algorithms. Therefore, we named these algorithms dropout k-means and dropout k-median.

Clustering algorithms, including k-means~\cite{steinhaus1956_kmeans,lloyd1982_kmeans} and k-median~\cite{charikar1999_kmedian,arora1998_kmedians}, are widely used in sensor placement~\cite{jain2010_kmeans}. Examples include relay node placement in wireless sensor networks~\cite{talgini2014_aerialnodeplacement,hajjar2017_relaynodeplacement} and temperature sensor placement in microprocessors~\cite{mukherjee2006_tempsensorplacement}. %There are, also, sensor placement strategies that do not rely on clustering~\cite{clouqueur2003_targetdetection, dhillon2003_coverage, zou2003_targetloc}.   

Prior work proposed a stochastic dropout k-means algorithm that applied dropout, a technique commonly used to regularize neural networks~\cite{srivastava2014_dropout}, to classic k-means clustering~\cite{zhang2016_dropoutkmeans}. In that paper, each iteration starts by a random dropout of some cluster centers, followed by an iteration of classic k-means on the remaining clusters. There is a convergence issue, especially when the number of clusters is small.

In contrast, in this paper, we modify the classic k-means and classic k-median objectives to consider all possible dropout outcomes, with weights given by the probability of occurrence of each outcome. While the number of different dropout outcomes is $2^K$, where $K$ is the number of clusters, we can effectively group outcomes and perform each cluster center update in polynomial time. We then position the marine buoys at the cluster centers obtained by the dropout k-means and dropout k-median algorithms.

Our contributions are:
\begin{itemize}[noitemsep,nolistsep]
    \item Formulate buoy placement as a clustering problem, where buoys, as cluster centers, may be disrupted. 
    \item Modify k-means and k-median with a dropout probability that models buoy disruption.
    \item Derive closed-form updates that considers all possible dropout outcomes in each iteration.
    \item Define a deterministic dropout k-means and a dropout k-median algorithm that runs in polynomial time, instead of exponential time, with the number of clusters.
\end{itemize}

The rest of the paper is structured as follows: Section~\ref{S:problemstatement} casts buoy placement as a clustering problem with the objective of minimizing of ship to buoy distances, with the probability of buoy dropout. In Section~\ref{S:method}, we describe dropout k-means and dropout k-median, and derive closed-form expressions for updating the centers and the clusters. Section~\ref{S:evaluation} compares the performance of dropout k-means and dropout k-median with classic k-means, classic k-median and stochastic dropout k-means implemented in prior work. Finally, Section~\ref{S:conclusion} concludes the paper.

\section{Problem Statement}\label{S:problemstatement}

We formulate buoy placement as a clustering problem where the goal is to place buoys, acting as cluster centers $\{c_k\}_{k=1}^K$, in order to maximize the probability of detecting ships, acting as data points $\{x_i\}_{i=1}^N$. The probability of detection $P_d$ is computed using the law of total probability:
\begin{equation}
        P_d = \sum_{\mathcal{V}\in\mathcal{M}}P(\mathcal{V})\sum_{\mathcal{S}\in\mathcal{P}(\mathcal{K})}P(\mathcal{S})\mathbf{1}_{\{\min_{i\in\mathcal{V},k\in\mathcal{S}}||x_i-c_k||\leq r\}},
        \label{eqn:P_d}
\end{equation}
where $\mathcal{M}$ is the set of ships, $\mathcal{V}$ is a set indices $i$ belonging to the same ship, and $P(\mathcal{V})$ is the probability of observing the ship; $\mathcal{K}=\{1,\dots,K\}$ is a set of cluster indices, $\mathcal{P}(\mathcal{K})$ is the power set of $\mathcal{K}$, that is, the set of all subsets of $\mathcal{K}$, with size $|\mathcal{P}(\mathcal{K})|=2^K$, representing all dropout outcomes, $\mathcal{S}$ is the set of remaining buoy indices in a particular dropout outcome, and $P(\mathcal{S})$ is the probability of the dropout outcome. The indicator function $\mathbf{1}$ represents ship detection. We assume detection of a ship to depend on the proximity of any of the ship's locations to any of the buoys remaining after dropout, where $r$ is the radius of detection. To simplify notation, we assume the probability of observing each ship to be the same, that is, $P(\mathcal{V})=\frac{1}{|\mathcal{M}|}, \forall \mathcal{V}\in\mathcal{M}$.  

The objective function given in equation~(\ref{eqn:P_d}) is piecewise constant and cannot be optimized via gradient based methods. Alternatively, we consider a convex relaxation of equation~(\ref{eqn:P_d}), which we minimize using a k-means type algorithm.

The classic k-means algorithm seeks to minimize the sum of square distances from data points to cluster centers, that is:
\begin{equation}
    \min_{c_1,\dots,c_K} \sum_{i=1}^N \min_{k\in\mathcal{K}} \|x_i-c_k\|^2,
    \label{eqn:kmeans_obj}
\end{equation}
where $\|\cdot\|$ is the Euclidean norm, also known as the $\mathit{L}^2$ norm.

On the other hand, the classic k-median algorithm seeks to minimize the sum of distances, that is:
\begin{equation}
    \min_{c_1,\dots,c_K} \sum_{i=1}^N \min_{k\in\mathcal{K}} \|x_i-c_k\|.
    \label{eqn:kmeans_obj}
\end{equation}

Since buoys may dropout as a result of disruption, changing the spatial configuration of the buoy network, we modify the classic k-means objective to include all possible dropout outcomes, given that a single buoy remains.

Thus, in dropout k-means, we add weights corresponding to the probability of dropout. Specifically, we minimize the objective function:
\begin{equation}
    \min_{c_1,\dots,c_K} \sum_{\mathcal{S}\in\mathcal{P}(\mathcal{K})\setminus\emptyset} P(\mathcal{S}) \sum_{i=1}^N \min_{k\in \mathcal{S}} \|x_i-c_k\|^2,
    \label{eqn:dkmeans_obj}
\end{equation}
where the probability of dropout outcome $\mathcal{S}$ is given as:
\begin{equation}
    P(\mathcal{S})=(1-p)^{|\mathcal{S}|}p^{K-|\mathcal{S}|},
\end{equation}
where dropout of each cluster center is assumed to be independent and identically distributed (i.i.d.) Bernoulli($p$). We do not consider the dropout outcome where no buoys remain, since, in that scenario, no ship can be detected, regardless of buoy placement. Thus, $P_d\leq1-P(\emptyset)=1-p^K$. 

Similarly, in dropout k-median, we add weights corresponding to the probability of dropout, and minimize the objective function:
\begin{equation}
    \min_{c_1,\dots,c_K} \sum_{\mathcal{S}\in\mathcal{P}(\mathcal{K})\setminus\emptyset} P(\mathcal{S}) \sum_{i=1}^N \min_{k\in \mathcal{S}} \|x_i-c_k\|.
    \label{eqn:dkmedian_obj}
\end{equation}

In dropout k-means and k-median, the terms $\min ||x_i-c_k||^2$ and $\min ||x_i-c_k||$ are considered as convex relaxations of the probability of no detection $\mathbf{1}_{\{ ||x_i-c_k||> r\}}$.

\section{Method}\label{S:method}

In this section, we describe dropout k-means, dropout k-median, and derive efficient implementations for their closed-form updates. Each update considers all possible dropout combinations. By first sorting cluster centers by their proximity to data points, each iteration scales proportionally to $NK\log(K)$, instead of $NK2^{K-1}$, where $N$ is the number of points and $K$ is the number of clusters. 

\subsection{Dropout k-means}
We first interchange the order of the summations in the dropout k-means objective shown in equation~(\ref{eqn:dkmeans_obj}): 
\begin{equation}
    \min_{c_1,\dots,c_K} \sum_{i=1}^N \sum_{\mathcal{S}\in\mathcal{P}(\mathcal{K})\setminus\emptyset} P(\mathcal{S})  \min_{k\in \mathcal{S}} \|x_i-c_k\|^2.
    \label{eqn:dkmeans_obj2}
\end{equation}

Fixing a data point $x_i$, we can calculate equation (\ref{eqn:dkmeans_obj2}) by grouping dropout outcomes. We denote a group of dropout outcomes, where the closest remaining cluster center to datapoint $x_i$ is cluster center $c_k$, as $\mathcal{S}_{ik}$. For each datapoint $x_i$, the groups of dropout outcomes are thus $\{S_{ik}\}_{k=1}^{K}$. We then show the calculation for the probability of a group of dropout outcomes $P(\mathcal{S}_{ik})$.

Let $\sigma_i:\mathcal{K}\to\mathcal{K}$ be a permutation such that the cluster centers $c_{\sigma_i(1)},\dots,c_{\sigma_i(K)}$ are sorted in increasing distances to $x_i$, such that $\|x_i-c_{\sigma_i(1)}\|<\dots<\|x_i-c_{\sigma_i(K)}\|$. Then, with probability $(1-p)$, $c_{\sigma_i(1)}$ is not dropped out, and remains the closest cluster center to $x_i$. For $c_{\sigma_i(2)}$ to be the closest cluster center to $x_i$, $c_{\sigma_i(1)}$ has to drop out and $c_{\sigma_i(2)}$ has to stay, which happens with probability $p(1-p)$. Using this reasoning, the probability that $c_{\sigma_i(j)}$ is the closest cluster center to data point $x_i$ is $p^{j-1}(1-p)$. On the other hand, the probability that $c_k$ is the closest cluster center to data point $x_i$ is $p^{\sigma_i^{-1}(k)-1}(1-p)$, where $\sigma_i^{-1}:\mathcal{K}\to\mathcal{K}$ is the inverse of $\sigma_i$, that is, $\sigma_i^{-1}(k)=j$ if $\sigma_i(j)=k$. The probability $P(\mathcal{S}_{ik})$ of the group of dropout outcomes $\mathcal{S}_{ik}$ is thus:

\begin{equation}
    P(\mathcal{S}_{ik})=p^{\sigma_i^{-1}(k)-1}(1-p),
    \label{eqn:dkmeans_obj_b}
\end{equation}

The dropout k-means objective from equation~(\ref{eqn:dkmeans_obj2}) is thus re-written as: 
\begin{equation}
\min_{c_1,\dots,c_K} \sum_{i=1}^N \sum_{k=1}^K P(\mathcal{S}_{ik}) \|x_i-c_{k}\|^2.
    \label{eqn:dropout_kmeans_sorted}
\end{equation}

We treat permutation $\sigma$ as a cluster assignment. For a fixed cluster assignment, a center $c_k$ is the minimizer of the weighted sum of square distances between itself and the points in that cluster, with weights given by the probabilities of each point being in that cluster. By taking a derivative with respect to $c_k$, we obtain that the center is a weighted average of points in that cluster:
\begin{equation}
c_k = \frac{\sum_{i=1}^N P(\mathcal{S}_{ik})x_i}{\sum_{i=1}^N P(\mathcal{S}_{ik})}.
\end{equation}

In both classic k-means and dropout k-means, the objective decreases monotonically. The algorithms are thus iterated to convergence, that is, when cluster assignments no longer change. The algorithm complexity scales as $NK\log(K)$, due to sorting the cluster centers, and is shown in Algorithm~\ref{alg:dkmeans}.

\begin{algorithm}
\caption{Dropout k-means}\label{alg:dkmeans}
 \KwData{Data points $\{x_i\}_{i=1}^N$}
 \KwResult{Cluster centers $\{c_k\}_{k=1}^K$}
 Initialize cluster centers over the support of data points
 \While{$\sigma\neq\sigma_{prev}$}{
  replace previous cluster assignment $\sigma_{prev}=\sigma$\;
  \textit{update cluster assignment:}\\
  \For{data point $x_i\in\{x_i\}_{i=1}^N$}{
  Sort $\{c_k\}_{k=1}^K$ by increasing distance to $x_i$\;
  Calculate assignment probabilities $\{P(\mathcal{S}_{ik})\}_{k=1}^K$\;
  }
  save cluster assignment $\sigma$\;
  \textit{update cluster centers:}\\
  \For{cluster center $c_k\in\{c_k\}_{k=1}^K$}{
  $c_k$ = $\sum_{i=1}^N P(\mathcal{S}_{ik})x_i/\sum_{i=1}^N P(\mathcal{S}_{ik})$.
  }}
\end{algorithm}

\subsection{Dropout k-median}

The dropout k-means objective is to minimize a weighted sum of square distances, and can be considered a convex relaxation of minimizing the probability of missed detection, where cluster centers are associated with the $\mathit{L}^2$ mean. Motivated by this convex relaxation approach, we also consider the algorithm minimizing a weighted sum of distances, where cluster centers are associated with the $\mathit{L}^1$ mean, also known as the geometric median. We denote this algorithm as dropout k-median. 

The minimizer $c_k$ is the geometric median instead of the geometric mean. Unfortunately, there is no closed-form solution for the geometric median. However, there is an iterative algorithm that converges to the geometric median, known as the Weiszfeld algorithm~\cite{weiszfeld1937geometricmedian}. We extend the Weiszfeld algorithm with weights given by equation (\ref{eqn:dkmeans_obj_b}) to obtain the cluster center update as:
\begin{equation}
c_k = \frac{\sum_{i=1}^N P(\mathcal{S}_{ik})\frac{x_i}{||x_i-c_k||}}{\sum_{i=1}^NP(\mathcal{S}_{ik})\frac{1}{||x_i-c_k||}}.
\end{equation}

\section{Evaluation}\label{S:evaluation}

We compare the ship detection probability of dropout k-means and dropout k-median with classic k-means, classic k-median and stochastic dropout k-means implemented in prior work~\cite{zhang2016_dropoutkmeans}, at the Gabonese Exclusive Economic Zone (EEZ) near West Africa. To simulate the passage of ships, we use AIS data. While AIS data may have inconsistencies from gaps in reporting, it is widely used in literature for visualizing fishing activity~\cite{kroodsma2018_gfw}, improving collision avoidance systems~\cite{zhang2015_nearmiss, gao2018_lstmrnn}, anomaly detection~\cite{laxhammer2008_anomalydet} and trade route identification~\cite{spiliopoulos2017_traderouteid}. 

\subsection{AIS Dataset}

 We downloaded AIS tracks from Global Fishing Watch's github repository~\cite{gfw_github}. Of the 1258 tracks in the repository, 55 tracks, with unequal contributions to a total of 313390 location reports, passed through Gabonese EEZ. The 313390 ship coordinates were then used for clustering.
 
%\begin{figure}
%    \centering
%    \includegraphics[width=.5\columnwidth]{gabon_AIStracks}
%    \caption{AIS tracks from fishing vessels that passed through the Gabonese Exclusive Economic Zone (EEZ).}
%    \label{fig:gabon_AIStracks}
%\end{figure}

\subsection{Evaluation Metrics}

We use ship detection probability ($P_d$), root mean square distance (RMSD), rate of convergence and runtime as metrics. 

The $P_d$ is computed as in equation~(\ref{eqn:P_d}), the RMSD is computed as:
\begin{equation}
    \text{RMSD} = \sqrt{\frac{\sum_{\mathcal{S}\in\mathcal{P}(\mathcal{K})\setminus\emptyset} P(\mathcal{S}) \sum_{i=1}^N \min_{k\in \mathcal{S}} \|x_i-c_k\|^2}{\sum_{\mathcal{S}\in\mathcal{P}(\mathcal{K})\setminus\emptyset} P(\mathcal{S}) \sum_{i=1}^N 1}}.
\end{equation}

Note that $P_d$ and RMSD may be calculated in polynomial time by grouping dropout outcomes, using the technique introduced in Section~\ref{S:method}. 

Rate of convergence is measured as number of iterations taken for the algorithm to converge, runtime is measured as the total time taken. With regard to convergence, classic k-means and dropout k-means have natural termination conditions, that is, when assignments no longer change. For classic k-median and dropout k-median, since the geometric median is found via an iterative algorithm, the algorithms might not have converged while the cluster assignments no longer change. However, when the assignments do not change, it means the solution is close to convergence. We thus take assignments not changing as the termination condition. For stochastic dropout k-means implemented in prior work~\cite{zhang2016_dropoutkmeans}, there is no natural termination condition. In addition, it has a convergence issue, especially when the number of clusters is small. For example, in the two clusters scenario, stochastic dropout k-means oscillates between one and two cluster configurations. For stochastic dropout k-means, we thus set a relaxed termination condition as all cluster centers moving by less than $\frac{r}{4}$, where $r$ is the sensor detection radius. 

\subsection{Results}

We evaluated the performance of deploying 5 buoys. We assume each buoy is disrupted with probability 0.3, since this is the fraction of buoys that missed reporting back to NDBC. We assume the 55 ships occur with the same probability and repeated the experiment 30 times. At the beginning of each experiment, we randomly generated cluster centers with k-means++ initialization~\cite{arthur2007_kmeanspp} and used the same initial set of cluster centers for all algorithms. The radius of detection $r$ was set at 10km, consistent with current sensor detection radii found in literature~\cite{fefilatyev2012_buoycam,prasad2017_buoycam}. The buoy arrangement computed by classic k-means, dropout k-means, stochastic dropout k-means~\cite{zhang2016_dropoutkmeans}, classic k-median and dropout k-median have ship detection probabilities of 38\%, 45\%, 45\%, 48\%, 52\%. The results, with mean and standard deviation from 30 trials, are summarized in Table~\ref{tab:results5and10}.
\begin{table}[h]
\centering
\caption{Performance comparison with K=5 clusters} \label{tab:results5and10}
\resizebox{\columnwidth}{!}{
\renewcommand{\arraystretch}{1.3}
\begin{tabular}{lcccc} \toprule
                 & number of iterations & total runtime (s) & RMSD (km) & $P_d$  \\ \midrule
classic k-means  & 31$\pm$12 & \textbf{0.5$\pm$0.1} & 154$\pm$9 & 0.38$\pm$0.03  \\ \midrule
dropout k-means  & \textbf{19$\pm$2} & 1.0$\pm$0.1 & \textbf{140$\pm$0} & 0.45$\pm$0.00\\ \midrule
stochastic dropout k-means~\cite{zhang2016_dropoutkmeans}       & 286$\pm$51 & 1.1$\pm$0.4  & 144$\pm$2 & 0.45$\pm$0.04\\ \midrule
classic k-median  & 54$\pm$18 & 1.2$\pm$0.4 & 151$\pm$7 & 0.48$\pm$0.05\\ \midrule
dropout k-median  & 79$\pm$22 & 4.2$\pm$0.9 & 141$\pm$1 &\textbf{0.52$\pm$0.03}\\ \bottomrule
\end{tabular}}
\end{table}

Both stochastic dropout k-means and dropout k-means have improved RMSD and $P_d$ over classic k-means. This is expected as dropout models buoy disruptions. In addition, stochastic dropout k-means should converge to the same result as dropout k-means in expectation, as observed. Stochastic dropout k-means, however, has a convergence issue, frequently hitting the limit of 300 iterations, despite having a more relaxed convergence condition. In addition, its RMSD and $P_d$ have larger variances than dropout k-means.

Dropout k-median also showed improved RMSD and $P_d$ over classic k-median. In addition, classic k-median and dropout k-median showed improved RMSD and $P_d$ over classic k-means and dropout k-means. This is because distance is a tighter upper bound than square distance on the probability of missed detection. In this experiment, minimizing a tighter upper bound produced a stronger algorithm. Lastly, with our efficient implementation, the runtimes of the dropout algorithms were comparable to the classic algorithms.

\section{Conclusion}\label{S:conclusion}

Dropout k-means and dropout k-median clustering give more robust buoy placement, where dropout aptly models buoy disruption. We proposed an efficient implementation for dropout k-means and extended the algorithm to dropout k-median, where distance is a tighter upper bound than square distance on the probability of missed detection. We simulated the placement of marine buoys at the cluster centers computed from ship AIS data in the Gabonese waters near West Africa. For 5 buoys, the ship detection probability of classic k-means, dropout k-means, classic k-median and dropout k-median are 38\%, 45\%, 48\% and 52\%.

\newlength{\bibitemsep}\setlength{\bibitemsep}{.2\baselineskip plus .05\baselineskip minus .05\baselineskip}
\newlength{\bibparskip}\setlength{\bibparskip}{0pt}
\let\oldthebibliography\thebibliography
\renewcommand\thebibliography[1]{%
  \oldthebibliography{#1}%
  \setlength{\parskip}{\bibitemsep}%
  \setlength{\itemsep}{\bibparskip}%
}

\bibliographystyle{IEEEbib}
\small{\bibliography{refs}}

\end{document}